\documentclass[
]{ceurart}

\usepackage{listings}
\usepackage{graphicx} % Required for inserting images

\usepackage{hyperref} % Optional: makes emails clickable
\usepackage{siunitx} % Required for the \num{} command

\usepackage{natbib}
\usepackage{pgfplots}
\usepackage{tikz}
\usetikzlibrary{backgrounds}
\usepackage{dirtree}
\usepackage{booktabs}
\usepackage{subcaption}
\usepackage{float}
\usepackage{amsmath}
\usepackage{amssymb}
\usepackage[T1]{fontenc}
\usepackage{graphicx}   % for \resizebox
\setcitestyle{square}
\usepgfplotslibrary{groupplots}
\pgfplotsset{compat=1.18}
\usepackage{cleveref} % Required for the \Cref{} command

\usetikzlibrary{positioning, calc, arrows.meta, shadows, decorations.pathreplacing}

\crefname{section}{Sect.}{Sects.}
\Crefname{section}{Section}{Sections}
\crefname{figure}{Fig.}{Figs.}
\Crefname{figure}{Figure}{Figures} 
\crefname{definition}{Definition}{Definitions}
\crefname{algorithm}{Algorithm}{Algorithms}
\crefname{equation}{}{}
\Crefname{equation}{Equation}{Equations}
\crefname{table}{Table}{Tables}
\Crefname{table}{Table}{Tables}
\crefname{reference}{Ref.}{Refs.} 
\Crefname{reference}{Reference}{References}

\crefformat{reference}{[#2#1#3]}
\Crefformat{reference}{Reference~[#2#1#3]}

\newcommand{\para}{\S} % Defines \para for paragraph symbol (§)
\newcommand{\Abs}{Abs.} % Defines \Abs for Absatz (Subsection) abbreviation
\newcommand{\Satz}{Satz}
\newcommand{\Nr}{Nr.}
\newcommand{\Buchst}{Buchst.}

\begin{document}

\copyrightyear{2026}
\copyrightclause{Copyright for this paper by its authors.
  Use permitted under Creative Commons License Attribution 4.0
  International (CC BY 4.0).}

\conference{
RAGE-KG 2026: The Second International Workshop on Retrieval-Augmented Generation Enabled by Knowledge Graphs, co-located
with ISWC 2026, October 25–26, 2026, Bari, Italy
}

\title{Verifiable Retrieval for Legal Reasoning through Structure-driven Knowledge Graph}

% \tnotetext[1]{You can use this document as the template for preparing your
%   publication. We recommend using the latest version of the ceurart style.}

%%
\author[1]{Ibrahim Siddig}[%
  % orcid=0000-0000-0000-0000,
  email=Ibrahim.Siddig@thi.de
]
\cormark[1]
\author[1]{Sviatoslav Tugeev}[%
  % orcid=0000-0000-0000-0000,
  email=Sviatoslav.Tugeev@thi.de
]
\author[1]{Munir Georges}[%
  % orcid=0000-0000-0000-0000,
  email=Munir.Georges@thi.de
]

\address[1]{Technische Hochschule Ingolstadt, Esplanade 10, 85049 Ingolstadt, Germany}

\cortext[1]{Corresponding author.}
\begin{abstract}
Any automation within reliability-critical contexts such as medicine or law has to be transparent and verifiable. Despite their advantages, current LLM-based systems do not meet these requirements due to their inherent black-box nature. Retrieval Augmented Generation (RAG) attempts to ground the generative process in the relevant interpretable documents, thereby reducing the risk of AI ’hallucinations’ and potentially enabling verification of the initial generation context. However, as the number and size of potential sources increases, naive retrieval approaches tend to perform worse, ignoring critical context elements. In this paper we aim to tackle this problem in the context of German Legal Code. Using structure-driven parsing to create a Knowledge Graph (KG) for the source document navigation and retrieval, we compare several KG-informed chunking and retrieval strategies and evaluate their effect on the accuracy of retrieved citations.
\end{abstract}


\maketitle

\section{Introduction}
\label{sec:intro}

The practical quality of legal answers is determined not only by their linguistic coherence and relevance to the context of the question, but by the degree to which they are supported by the relevant legal documents. This support is usually demonstrated by citing specific provisions, enabling the end user to verify legal validity of the answer. In practice, a legal text that is linguistically correct but cites the wrong sources or cites them incorrectly is not just useless, but harmful, as it looks trustworthy while being factually wrong or incomplete. This is especially relevant in the context of AI systems, which are already known for their ability to produce false but extremely plausible results \cite{ji2023survey}. 
In the context of RAG, where the output is generated based on the relevant law texts, this problem can be separated into three key issues:
\begin{itemize}
    \item Whether all source documents or their segments which contain information necessary for a valid legal judgment are retrieved based on the initial query (factual description of the case).
    \item Whether the generated answer corresponds to the facts of the case and the content of the retrieved documents.
    \item Whether the retrieved documents are cited correctly within the final answer.
\end{itemize}
For German law, key prerequisites for studying this topic have been available for some time. The OpenLegalData corpus \footnote{\url{https://huggingface.co/datasets/harshildarji/openlegaldata}} \cite{ostendorff2020towards, darji2025segmentation} provides \num{424000} German court decisions with structured metadata, separated into the following sections: \textbf{Tenor} which explains the final decision, \textbf{Tatbestand} which shows statements of case facts; and \textbf{Entscheidungsgründe} which states the reasoning behind the final case decision. This extensive data is, however, insufficient, without a reliable way of processing it. This paper seeks to address the first issue in the RAG pipeline by exploring the viability of a structure-driven knowledge graph for addressing accuracy and token efficiency of retrieval in legal AI. In order to do that, we attempt to answer the following research questions:

\begin{enumerate}
    \item \textbf{RQ1:} How does a structure-driven KG influence the retrieval relevance and citation precision in the RAG pipeline for legal documents?
    \item \textbf{RQ2:} What level of granularity in structure-aware text chunking is most beneficial for the relevance of retrieved sources and the overall retrieval efficiency?
    \item \textbf{RQ3:} How do different one-hop graph-driven chunk expansion strategies affect retrieval accuracy across different chunking levels?
\end{enumerate}

\subsection{Key contributions}
\label{subsec:contributions}

\begin{enumerate}
    \item \textbf{A knowledge graph of German federal statute law.} 
    \num{58} codes, an eight-level typed hierarchy from root down to \textit{Unterbuchstabe} which represents the atomic level of the German federal law, \num{168000} nodes, and \num{240000} edges of which \num{22000} are intra-code cross-references derived from in-statute citation tagging. The knowledge graph covers \num{90.3}\% of the ground truth citations in the OpenLegalData BGH evaluation set. The remaining $9.7\%$ are out-of-scope or unretrievable.
    
    \item \textbf{A structured token-normalized citation-faithfulness metric.} We evaluate retrieved citations using full hierarchical path matching (\texttt{Buch / \para\ / \Abs\ / \Satz\ / \Nr\ / \Buchst})\footnote{\textbf{\texttt{Buch}}: Book/Code; \textbf{\texttt{\para}}: Section (\textit{Paragraph}); \textbf{\texttt{\Abs}}: Subsection (\textit{Absatz}); \textbf{\texttt{\Satz}}: Sentence; \textbf{\textit{\texttt{\Nr}}}: Number (\textit{Nummer}); \textbf{\texttt{\Buchst}}: Letter (\textit{Buchstabe}).}. Leveraging the KG's hierarchy \cref{fig:kg-hierarchy}, the metric evaluates accuracy depth-wise to distinguish granular errors, such as citing an incorrect sentence within a paragraph versus citing an entirely wrong legal code. To decouple evaluation from chunk size bias, we normalize recall against total retrieved context length \cref{eq:toer}.
      
    \item \textbf{Detailed comparison of chunking and one-hop retrieval strategies.} On a held-out set of \num{12037} BGH cases, we compare different levels of granularity in law text chunking and one-hop KG-based retrieval strategies using the proposed metrics. 
\end{enumerate}

\section{Related Work}
\label{sec:related}
\cite{ostendorff2020towards} released \num{250000} German
court decisions with structured metadata; the subsequent segmentation work of ~\cite{darji2025segmentation} released a cleaned JSONL corpus of
\num{251038} decisions with Tenor / Tatbestand / Entscheidungsgründe splits. The \textit{Bundesrecht} library \cite{darji2026bundesrecht} covers the upstream task of parsing German statutory reference strings, providing a structured JSONL corpus of \num{6873} federal laws. Our knowledge graph is complementary rather than competing. While \textit{Bundesrecht} optimizes for citation-string parsing and depth-aware lookup over a flat representation, we build a graph with defined cross-reference edges.

\citep{de2025graph} proposed a hierarchical and
temporal graph-RAG architecture for legal norms but did not evaluate on
real court decisions. \cite{nigam2025nyayarag} presented NyayaRAG, a
statute-plus-case RAG framework for Indian legal judgment prediction; their
evaluation conflates flat statute retrieval and case retrieval. \cite{ongris2025benchmarking} at RAGE-KG 2025 benchmarked KG-based RAG systems on legal documents. Our work differs in scoping the retrieval knowledge base
to statute law alone, holding case law out entirely as evaluation gold, and
measuring citation faithfulness at the level courts actually cite.

Prior evaluations of legal citation accuracy typically match at a flat or abstract
level~\cite{chen2026legalcitebench, guha2023legalbench}, which cannot
distinguish \para~574~\Abs~2~\Nr~2 from \para~574~\Abs~3. Our structured
metric closes this gap by accounting for strict hierarchical subdivisions critical for legal documents. Empirical evaluations demonstrate that language models fail to robustly utilize long inputs, exhibiting severe performance degradation when critical information is positioned in the middle of the context window \cite{liu2024lost}. Modern frameworks increasingly employ multi-level hierarchical segmentation, ensuring semantic continuity across diverse text scales while preserving structural layout \cite{lu2026hichunk}. The necessity of token-efficient graph retrieval is further highlighted by \textsc{TERAG} \cite{xiao2025terag}, which adapts Personalized PageRank \cite{gutierrez2024hipporag} to cut GraphRAG token costs down to $3\%$--$11\%$ while maintaining retrieval fidelity. In a similar spirit of scaling graph-based retrieval to large legal corpora, our work demonstrates that optimizing node granularity directly tackles this token bottleneck, delivering high recall efficiency per token.

\section{Methodology}

\subsection{Knowledge graph construction}

To build our KG, a set of \num{58} codes was compiled based on the citations in the
\textit{Entscheidungsgründe} and law references sections of the OpenLegalData corpus. The
full hierarchy, from the code down to the \emph{Buchstabe}, is extracted from the official
consolidated text published at \textit{gesetze-im-internet.de} using our custom regex parser,
established in \cite{Siddig2025_1260} and expanded here with additional patterns to enable
processing of documents with more diverse structures. Citations occurring within a statute
are tagged during parsing and represented as directed \texttt{cites\_*} edges; all
$22{,}253$ of them resolve within a single code, so the result is a forest of \num{58}
disjoint code graphs. The resulting KG in \cref{fig:kg-hierarchy} includes the following
hierarchical node types, representing the structure of the original document:
\textit{\textbf{Buch $\cdot$ Titel $\cdot$ \S{} $\cdot$ Absatz $\cdot$ Satz $\cdot$
Nummer $\cdot$ Buchstabe}} \footnote{\textbf{\textit{Buch}}: Book/Code; \textbf{\textit{Titel}}: Title;
\textbf{\textit{\S{}}}: Section (\textit{Paragraph}); \textbf{\textit{Absatz}}: Subsection;
\textbf{\textit{Satz}}: Sentence; \textbf{\textit{Nummer}}: Number;
\textbf{\textit{Buchstabe}}: Letter.}

\definecolor{citeblue}{HTML}{1B6FB3}

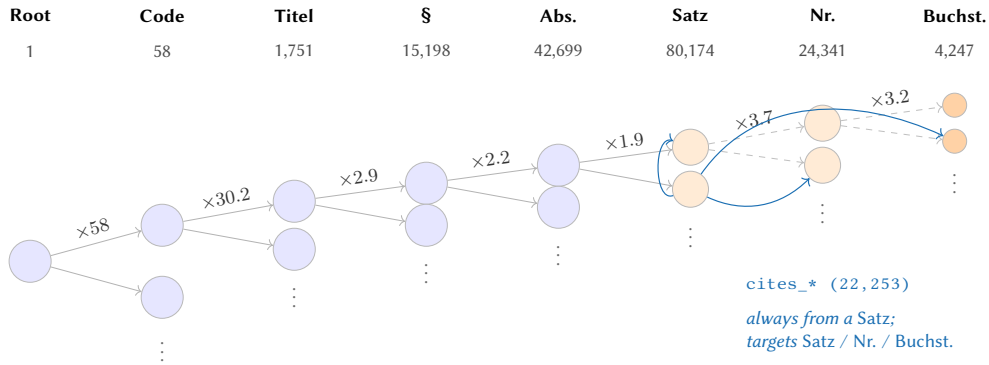
\begin{figure}[h]
\centering
\resizebox{\linewidth-80pt}{!}{%
\begin{tikzpicture}[
  inode/.style={circle, draw=gray!55, fill=blue!10, minimum size=7mm, inner sep=0pt},
  onode/.style={circle, draw=gray!55, fill=orange!16, minimum size=6mm, inner sep=0pt},
  leaf/.style={circle, draw=gray!55, fill=orange!35, minimum size=4mm, inner sep=0pt},
  e/.style={->, draw=gray!60},
  eopt/.style={->, draw=gray!50, dashed},
  % cites_* must NOT reuse the dashed grey of eopt: the optional-containment edges
  % (Satz->Nr, Nr->Buchst) are already dashed grey, so a dashed grey cross-reference is
  % indistinguishable from them. Own colour, solid, slightly heavier.
  cites/.style={->, draw=citeblue, line width=0.55pt},
fan/.style= {font=\footnotesize, text=black!80, midway, above, sloped},
hdr/.style= {font=\footnotesize\bfseries, text=black, align=center},
cnt/.style= {font=\footnotesize, text=black!70, align=center},
note/.style={font=\footnotesize\itshape, text=black!70, align=center},
dots/.style={font=\footnotesize, text=black!60}
]

% ---------- nodes ----------
\node[inode] (root) at (  0mm,  0mm) {};

\node[inode] (c1)   at ( 22mm,  6mm) {};
\node[inode] (c2)   at ( 22mm, -6mm) {};
\node[dots]         at ( 22mm,-14mm) {$\vdots$};

\node[inode] (k1)   at ( 44mm, 10mm) {};
\node[inode] (k2)   at ( 44mm,  2mm) {};
\node[dots]         at ( 44mm, -5mm) {$\vdots$};

\node[inode] (p1)   at ( 66mm, 13mm) {};
\node[inode] (p2)   at ( 66mm,  6mm) {};
\node[dots]         at ( 66mm, -1mm) {$\vdots$};

\node[inode] (u1)   at ( 88mm, 16mm) {};
\node[inode] (u2)   at ( 88mm,  9mm) {};
\node[dots]         at ( 88mm,  2mm) {$\vdots$};

\node[onode] (s1)   at (110mm, 19mm) {};
\node[onode] (s2)   at (110mm, 12mm) {};
\node[dots]         at (110mm,  5mm) {$\vdots$};

\node[onode] (m1)   at (132mm, 23mm) {};
\node[onode] (m2)   at (132mm, 16mm) {};
\node[dots]         at (132mm,  9mm) {$\vdots$};

\node[leaf]  (b1)   at (154mm, 26mm) {};
\node[leaf]  (b2)   at (154mm, 20mm) {};
\node[dots]         at (154mm, 14mm) {$\vdots$};

% ---------- containment edges ----------
\draw[e] (root) -- node[fan]{$\times 58$}   (c1);
\draw[e] (root) -- (c2);
\draw[e] (c1)   -- node[fan]{$\times 30.2$} (k1);
\draw[e] (c1)   -- (k2);
\draw[e] (k1)   -- node[fan]{$\times 2.9$}  (p1);
\draw[e] (k1)   -- (p2);
\draw[e] (p1)   -- node[fan]{$\times 2.2$}  (u1);
\draw[e] (p1)   -- (u2);
\draw[e] (u1)   -- node[fan]{$\times 1.9$}  (s1);
\draw[e] (u1)   -- (s2);

% enumeration levels are OPTIONAL (sparse) -> dashed
\draw[eopt] (s1) -- node[fan]{$\times 3.7$} (m1);
\draw[eopt] (s1) -- (m2);
\draw[eopt] (m1) -- node[fan]{$\times 3.2$} (b1);
\draw[eopt] (m1) -- (b2);

% ---------- cross-reference edges (within-code) ----------
% All three leave the SAME Satz node: cites_* always originates at a Satz, and targets a
% Satz (15,030), a Nummer (6,805) or a Buchstabe (418). Drawing only the Satz->Nummer case,
% as the previous version did, reads as though a Satz can cite nothing else.
\draw[cites] (s2) to[out=200,in=160] (s1);                % Satz -> Satz       (68%)
\draw[cites] (s2) to[out=-25,in=-135] (m2);               % Satz -> Nummer     (31%)
\draw[cites] (s2) to[out=55,in=155] (b2);                 % Satz -> Buchstabe  ( 2%)
\node[font=\footnotesize\ttfamily, text=citeblue, anchor=west] at (118mm, -4mm) {cites\_* (22{,}253)};
\node[note, text=citeblue, anchor=north west, align=left] at (118mm, -7mm)
     {always from a \emph{Satz};\\targets \emph{Satz} / \emph{Nr.} / \emph{Buchst.}};

% ---------- column headers and node counts ----------
\node[hdr] at (0mm,  41mm) {Root};       \node[cnt] at (0mm,  35mm) {1};
\node[hdr] at (22mm, 41mm) {Code};       \node[cnt] at (22mm, 35mm) {58};
\node[hdr] at (44mm, 41mm) {Titel};      \node[cnt] at (44mm, 35mm) {1{,}751};
\node[hdr] at (66mm, 41mm) {\S{}};       \node[cnt] at (66mm, 35mm) {15{,}198};
\node[hdr] at (88mm, 41mm) {Abs.};      \node[cnt] at (88mm, 35mm) {42{,}699};
\node[hdr] at (110mm,41mm) {Satz};      \node[cnt] at (110mm, 35mm) {80{,}174};
\node[hdr] at (132mm,41mm) {Nr.};       \node[cnt] at (132mm, 35mm) {24{,}341};
\node[hdr] at (154mm,41mm) {Buchst.};   \node[cnt] at (154mm, 35mm) {4{,}247};
\end{tikzpicture}%
}

\caption{\textbf{The Statute knowledge graph hierarchy} includes $58$ federal codes, $168{,}469$ nodes, and $240{,}507$ edges. Column headers display each level's name and total node count, noting that not all \textit{Sätze} contain a \textit{Nummer} or \textit{Buchstabe}. Edge labels provide the average number of child nodes per branching parent at that level. Cross-reference edges marked \texttt{cites\_*}, drawn in blue. Citations originate at a \emph{Satz} node since they are tagged in sentence text and target a \emph{Satz} ($15{,}030$), \emph{Nummer} ($6{,}805$) or \emph{Buchstabe} ($418$) node within the same code. The final graph is a forest of 58 disjoint code graphs, connected to the empty parent root node.}

\label{fig:kg-hierarchy}
\end{figure}

Each node is linked to the associated segment of text. The parent node contains the text of all its child nodes, i.e. if an \textit{Absatz} node has multiple \textit{Satz} child nodes, text associated with the \textit{Absatz} node includes text of all the child \textit{Satz} nodes. The constructed KG consists of the single root node not related to any specific document, which then diverges into multiple subtrees, each representing specific legal code. While there are no connecting edges between the subtrees, the decision to unify them into a single graph was guided by the fact that in the OpenLegalData corpus multiple codes are regularly cited together, demonstrating that legal decisions require access to an arbitrary number of codes simultaneously. Unlike the framework described in \cite{de2025graph}, the resulting KG includes only two types of edges: \texttt{contains\_*}, to represent the hierarchical structure of the document, and \texttt{cites\_*}, to represent citation references. The temporal dimension, usually expressed via \texttt{supersedes\_*} or \texttt{amends\_*} is not relevant to the selected documents.

\subsection{Experiment setup}
All text segments, including statutory chunks and query fields, are mapped to a continuous vector space using \texttt{KaLM-Embedding-Gemma3-12B-2511} \cite{zhao2025kalm}. This 12-billion parameter text embedding and retrieval model is optimized for high-density retrieval and achieves state-of-the-art multilingual performance on benchmarks such as MMTEB. 

\begin{figure}[ht]
    \centering
    \includegraphics[width=\linewidth-80pt]{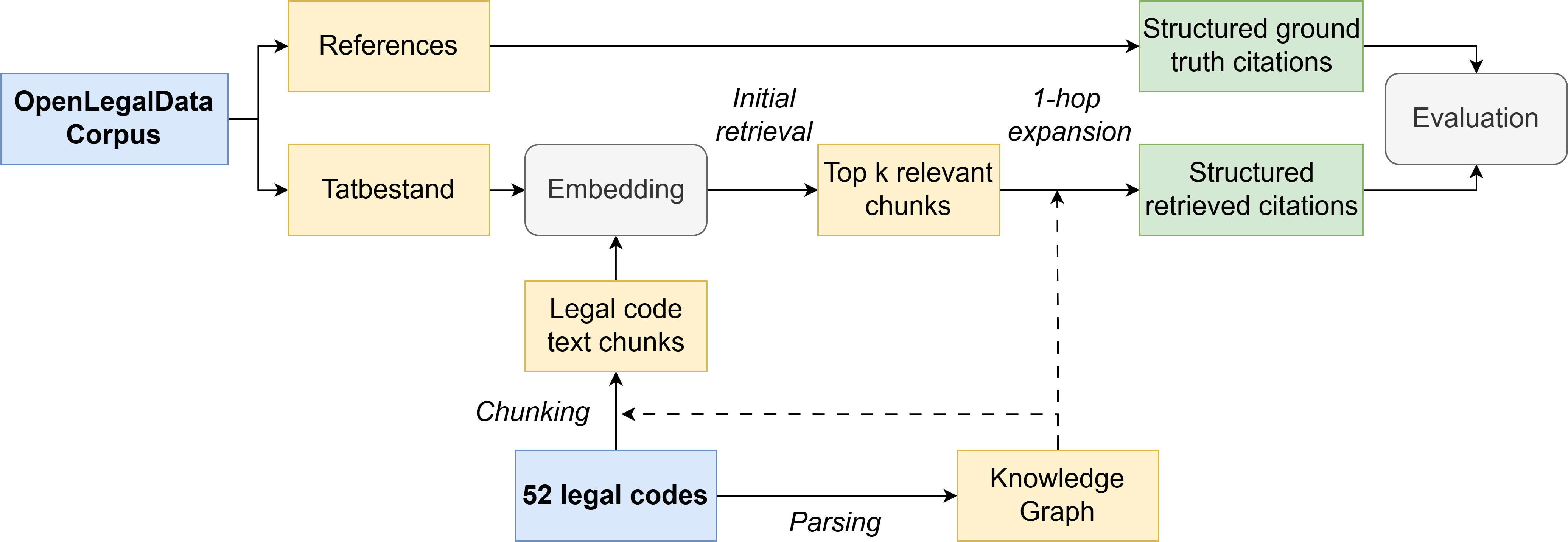}
    \caption{After the initial chunking, whether naive or KG-driven, the resulting chunks are embedded using KaLM-Embedding-Gemma3-12B. For each entry in the OpenLegalData corpus, the \emph{Tatbestand} section, which describes the facts of the case, is embedded the same way to serve as pipeline input, and the top-\textit{k} relevant chunks of the legal codes are retrieved. This list is then optionally expanded through a 1-hop expansion strategy using KG. The final structured list of retrieved references is compiled based on the IDs of associated nodes in KG and compared to the structured list of references from the corresponding section of the same corpus entry.}
    \label{fig:eval_pipeline}
\end{figure}

To evaluate the impact of different chunking granularities and multi-hop graph retrieval strategies, we establish the pipeline illustrated in \cref{fig:eval_pipeline}.

\subsubsection{Chunking strategies}
Structure-driven KG enables granular chunking of the original document, selecting chunks based on different node levels. Different chunking sizes offer a tradeoff between the probability that any retrieved chunk will include relevant information and the increasing context window, coupled with a reduced effect of any single chunk on the final model output. To compare different chunking strategies for processing legal documents using KG, in this paper we explored the effectiveness of chunking on the following levels, listed in increasing level of granularity:
\textbf{\S{} $\cdot$ Absatz $\cdot$ Satz $\cdot$ Nummer $\cdot$ Buchstabe}\footnote{\textit{\textbf{\S{}}}: Section (\textit{Paragraph}); \textbf{\textit{Absatz}}: Subsection; \textbf{\textit{Satz}}: Sentence; \textbf{\textit{Nummer}}: Number; \textbf{\textit{Buchstabe}}: Letter.}

These levels were compared to each other and against a naive chunking approach, which did not utilize KG and simply chunked the texts of the legal codes into consecutive 100-word chunks, ignoring the document structure. The constant 100-w size was chosen to ensure structure-agnostic approach.

\subsubsection{One-hop retrieval strategies}
\label{subsec:1-hop_expansion}
As each text chunk is linked to a specific node, a set of relevant text chunks can be further expanded using KG edges. As KG includes two types of edges, we evaluated three different one-hop expansion strategies:
\begin{itemize}
    \item \textbf{Parent node.} This strategy utilized only the \texttt{contains\_*} edge type, representing hierarchical links in KG. As a result, this expansion included the texts of parent nodes of the initially retrieved top-\textit{k}.
    \item \textbf{Citation.} This strategy utilized only the \texttt{cites\_*} edge type, representing reference links in KG, including the nodes cited directly in the initially retrieved top-\textit{k}. 
    \item \textbf{Hybrid.} This strategy utilized both types of links at the same time.
\end{itemize}
These strategies were compared to the baseline where only the originally retrieved top-\textit{k} nodes were used.

\subsubsection{Citation Matching and Evaluation Metric}
\label{sec:eval}

Since the paper is focused on a retrieval-only segment of RAG pipeline, we compare the  set of retrieved citations contained in the top-$k$ chunks (both initial and expanded) against the case's ground truth citations. Both sets are normalized into an ordered tuple over the German statutory hierarchy to ensure consistent comparison: \textit{Buch\footnote{\textit{\textbf{Buch}}: Book/Code.} $\cdot$ \S{} $\cdot$ Absatz $\cdot$ Satz $\cdot$ Nummer $\cdot$ Buchstabe}

Ground truth citations are parsed from the case references, while retrieved citations are decoded directly from the KG IDs of the retrieved nodes. To evaluate granular citation correctness, we report depth-stratified evaluation using a \emph{prefix match rule}, where agreement is credited as long as all levels explicitly specified in both the retrieved and ground truth citations match. Unspecified deeper levels such as a missing \textit{Nummer} or \textit{Buchstabe} are counted as implicit hits. This mirros the naturally heterogeneous judicial citation practices. For detailed statistics refer to  \cref{tab:depth_stratified_toer}.

Because raw recall (R) is monotonic in chunk size, as bigger chunks trivially cover more
citations at a massive token cost, we instead apply a custom metric, \textbf{token
efficiency recall (ToER) per $1{,}000$ words of chunk length} as defined in
\cref{eq:toer}. Standard F1 score is insufficient, since precision falls as recall rises and the two
effects cancel each other, reducing the  score sensitivity to the extent of added context. To isolate retrieval quality from coverage constraints in
\cref{subsec:contributions}, the ground truth is restricted to citations whose codes exist
in the constructed knowledge graph.

\begin{equation}
\mathrm{ToER} = \frac{10^{3}}{W_{\ell}} \cdot
\underbrace{\frac{1}{|\mathcal{G}|}\sum_{C_g \in \mathcal{G}} \max_{C_r \in \mathcal{R}_k} \mathrm{Match}(C_r, C_g)}_{\text{prefix recall}}
\label{eq:toer}
\end{equation}

\noindent
Where $W_{\ell}$ is the mean length of a single chunk at the
granularity $\ell$ averaged over the \emph{entire set of chunks in the whole corpus} at that
granularity. It is thus a
property of the chunking scheme rather than of any individual retrieval, and is constant across
queries, across $k$, and across expansion strategies. $\mathcal{R}_k$ is the set of citations decoded
from the retrieved nodes, $\mathcal{G}$ denotes the set of ground truth citations for the case and $\mathrm{Match}(C_r, C_g) \in \{0,1\}$ evaluates the
prefix-rule agreement between a retrieved citation $C_r$ and a ground truth citation $C_g$. Both citations are truncated at the evaluation depth being reported. \Cref{tab:depth_stratified_toer} presents the depth-stratified ToER performance. Reported values are micro-averaged across $12{,}037$ cases.

\section{Results \& Discussion}
\label{sec:results_discussion}
As shown in \cref{tab:depth_stratified_toer}, structure-aware chunking strategy with higher granularity consistently dominates across all evaluation depths. Specifically, the \textbf{Nummer} chunking granularity achieves the highest ToER at the \textit{Code} (14.23) and \textit{\S{}} (3.25) evaluation depths. Meaning that this approach retrieves the most relevant chunks of the source texts at the lowest token cost, when evaluated up to this depth. Chunking at \textbf{Satz} level yields the strongest performance at every deeper evaluation depth, reaching 2.28 at \textit{Absatz}, 1.53 at \textit{Satz}, and 2.79 at \textit{Nummer}. Both configurations recover approximately three times more citations per 1,000 words than the \textbf{Naive-100w} chunking baseline, showing  14.23 vs.\ 4.19 at \textit{Code}, 3.25 vs.\ 1.10 at \textit{\S{}}, and 2.79 vs.\ 0.81 at \textit{Nummer} evaluation depths. 

Notably, this performance increase is driven by the chunking granularity itself rather than merely introducing structural awareness. Chunking at higher levels, such as \textbf{\S{} (Paragraph)}, underperforms, even falling below the unstructured \textbf{Naive-100w} baseline with a ToER of only 3.06 at \textit{Code} and 0.85 at \textit{\S{}} evaluation depths. Chunking at \textbf{Absatz} then represents an intermediate transition (7.78 and 1.93, respectively). Across all chunking granularities, ToER drops abruptly between the \textit{Code} and \textit{\S{}} evaluation depth, such as \textbf{Nummer} falling from $14.23 \rightarrow 3.25$. This sharp drop exposes the heavy efficiency penalty the approach pays when moving from broad, code-level retrieval down to precise citations.

\begin{table}[h]
\centering
\small
\begin{tabular}{lcccccc}
\hline
 & \multicolumn{6}{c}{\textbf{Evaluation depth (strict match up to) [ToER $\uparrow$]}} \\ \cline{2-7} 
\textbf{Granularity (Chunks)} & \textit{Code} & \textit{§} & \textit{Absatz} & \textit{Satz} & \textit{Nummer} & \textit{Buchst}\\ \hline

\textbf{Buchstabe} & 12.89 & 2.93 & 2.05 & 1.35 & 1.91 & 0.50 \\
\textbf{Nummer} & \textbf{14.23} & \textbf{3.25} & 2.27 & 1.49 & 2.17 & 0.28 \\
\textbf{Satz} & 13.59 & 3.20 & \textbf{2.28} & \textbf{1.53} & \textbf{2.79} & \textbf{0.52} \\
\textbf{Absatz} & 7.78 & 1.93 & 1.36 & 1.37 & 1.57 & 0.30 \\
§ (\textbf{Paragraph}) & 3.06 & 0.85 & 0.92 & 0.90 & 0.94 & 0.16 \\ 
\textbf{Naive-100w (baseline)} & 4.19 & 1.10 & 0.85 & 0.72 & 0.81 & 0.30 \\
\hline
\end{tabular}
\caption{Depth-stratified ToER at top $k=5$ retrieved chunks. Structure-aware chunking granularities (\textbf{Nummer} and \textbf{Satz}) yield the highest scores across all depths, outperforming both the \textbf{Naive-100w baseline} and \textbf{paragraph-level (\S{})} chunking strategies. Values in bold indicate the best-performing chunking granularity per evaluation depth.}
\label{tab:depth_stratified_toer}
\end{table}

In \cref{fig:expansion-tokeneff}, all evaluations are conducted at the \textit{Absatz} evaluation depth, which is the depth at which ground truth citations most often terminate. It displays the effect of taking into account the token cost of context expansion. \Cref{fig:exp-uncharged} measures the ToER each strategy contributes without taking into account the token cost.
The three expansion strategies discussed in \cref{subsec:1-hop_expansion} behave
differently. The \texttt{cited} expansion strategy yields a small, uniform ToER gain across all chunking granularity levels, raising ToER $2.05 \rightarrow 2.17$ at \textbf{Buchstabe}, $2.27 \rightarrow 2.40$ at \textbf{Nummer}, and $2.28 \rightarrow 2.40$ at
\textbf{Satz} chunking granularities. With the \texttt{parent} strategy, ToER score remains fixed across most chunking levels, except at \textbf{Absatz}, where the direct parent represents the entire statutory paragraph (\S{}) and ToER increases from 1.36 to 2.13. At all chunking granularity levels except \textbf{Absatz}, \texttt{parent} expansion reproduces the baseline values exactly (2.05, 2.27, 2.28, and 0.92, respectively). Combining both expansion approaches in the \texttt{Hybrid} expansion further lifts the ToER score of the \textbf{Absatz} chunking granularity to 2.18. 

Subsequently, context expansion cannot resolve a poor initial choice of chunking size. Even when we apply
our most generous expansion (\texttt{Hybrid}) to the \textbf{Absatz} chunking depth, its performance
($2.18$) still falls short of smaller, \texttt{un-expanded}  \textbf{Nummer} ($2.27$) or
\textbf{Satz} ($2.28$) chunking depths. When a set of chunks retrieved after using these chunking strategies is expanded with the \texttt{Hybrid} approach, they both reach ToER $=2.40$. 

\Cref{fig:exp-charged} takes the token cost of context expansion into account. It replaces $W_{\ell}$ with $W_{\ell}^{m}$, denoting the average total words delivered per chunk, which in this case means the base text plus appended hop context, averaged across the entire corpus. Every expansion strategy then falls below its own unexpanded baseline, and the largest apparent gain inverts: \textbf{Absatz}~+~\texttt{parent} moves the ToER $1.36 \rightarrow 0.36$, because adding \S{} text raises the delivered context from $70.6$ to $415.0$ words. These
findings reveal that the initial choice of chunking granularity remains the dominant factor in retrieval efficiency under either configuration.

% \Cref{fig:exp-charged} takes the token cost of context expansion into account, setting $W$ to the mean number of context words per  chunk at the chosen chunking depth, taking into account the context expansion for the retrieved chunks, rather than the mean length of the chunks alone.

% (\cref{app:charged-toer})

%%%%%%%%%%%%%%%%%%%%%%%%%%%%%%%%%%%%%%%%%%%%%%%%%%%%%%%%%%%%%%%%%%%%%%%%%%%%%

\definecolor{expbase}{HTML}{B0AEA6}
\definecolor{expcited}{HTML}{1B9E77}
\definecolor{expparent}{HTML}{E6741F}
\definecolor{expHybrid}{HTML}{2A78D6}

% Shared panel style
\pgfplotsset{
  exppanel/.style={
    ybar,
    width=\linewidth, height=5.9cm,
    bar width=4pt,
    clip=false,
    ymin=0, ymax=2.6,
    ytick={0,0.5,1.0,1.5,2.0,2.5},
    ymajorgrids, grid style={gray!18, thin},
    axis lines=left,
    enlarge x limits=0.15,
    axis line style={gray!55},
    tick style={gray!55},
    every axis plot/.append style={draw=black!45, line width=0.35pt},
    symbolic x coords={Buchstabe,Nummer,Satz,Absatz,Paragraph},
    xtick=data,
    xticklabels={Buchst.,Nummer,Satz,Absatz,\S},
    x tick label style={font=\scriptsize, rotate=30, anchor=north east},
    tick label style={font=\scriptsize},
    ylabel style={font=\scriptsize},
  },
}

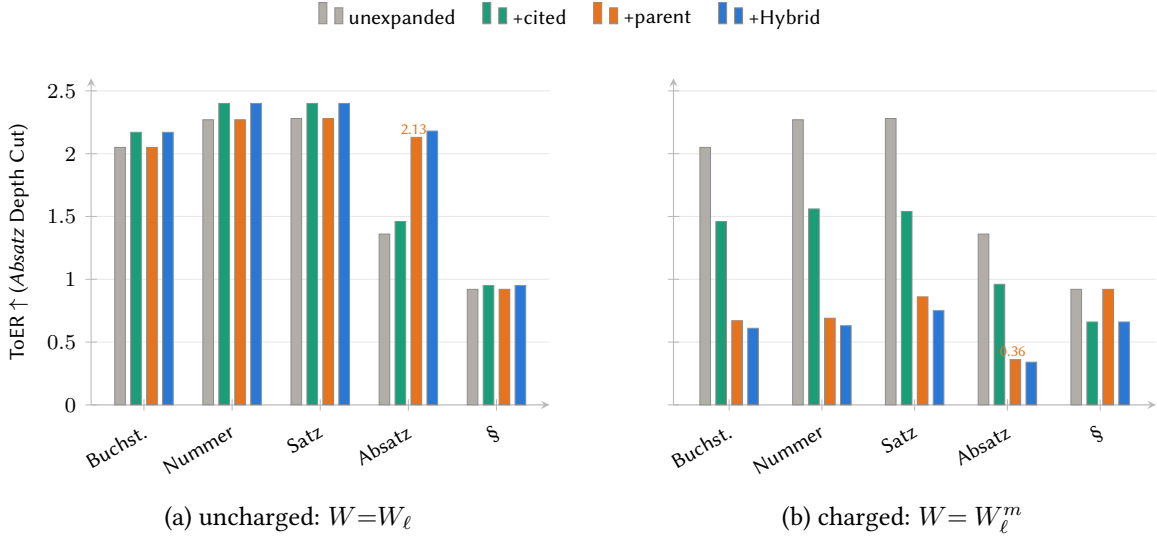
\begin{figure}[t]
\centering
% ---- (a) expansion uncharged -----------------------------------------------
\begin{subfigure}[t]{0.48\linewidth}
\centering
\begin{tikzpicture}
\begin{axis}[exppanel,
    ylabel={ToER $\uparrow$ (\textit{Absatz} Depth Cut)},
    % legend spans both panels; nudge the 1.05 if it sits off-centre
    legend style={
        at={(1.05,1.12)}, anchor=south, legend columns=4, draw=none,
        font=\scriptsize,
        /tikz/every even column/.append style={column sep=8pt}
    },
]
\addplot[fill=expbase]   coordinates {(Buchstabe,2.05)(Nummer,2.27)(Satz,2.28)(Absatz,1.36)(Paragraph,0.92)};
\addplot[fill=expcited]  coordinates {(Buchstabe,2.17)(Nummer,2.40)(Satz,2.40)(Absatz,1.46)(Paragraph,0.95)};
\addplot[fill=expparent] coordinates {(Buchstabe,2.05)(Nummer,2.27)(Satz,2.28)(Absatz,2.13)(Paragraph,0.92)};
\addplot[fill=expHybrid] coordinates {(Buchstabe,2.17)(Nummer,2.40)(Satz,2.40)(Absatz,2.18)(Paragraph,0.95)};
\legend{unexpanded, +cited, +parent, +Hybrid}

% Dashed baselines
 ([xshift=-7pt]axis cs:Buchstabe,2.05) -- ([xshift=7pt]axis cs:Buchstabe,2.05);
 ([xshift=-7pt]axis cs:Nummer,2.27)    -- ([xshift=7pt]axis cs:Nummer,2.27);
 ([xshift=-7pt]axis cs:Satz,2.28)      -- ([xshift=7pt]axis cs:Satz,2.28);
 ([xshift=-7pt]axis cs:Absatz,1.36)    -- ([xshift=7pt]axis cs:Absatz,1.36);
 ([xshift=-7pt]axis cs:Paragraph,0.92) -- ([xshift=7pt]axis cs:Paragraph,0.92);

\node[anchor=south, font=\tiny, text=expparent, inner sep=1pt]
      at ([xshift=2pt]axis cs:Absatz,2.13) {2.13};
\end{axis}
\end{tikzpicture}
% \vspace{-4pt}
\caption{uncharged: $W{=}W_{\ell}$}
\label{fig:exp-uncharged}
\end{subfigure}%
\hfill
% ---- (b) expansion charged -------------------------------------------------
\begin{subfigure}[t]{0.5\linewidth}
\centering
\begin{tikzpicture}
\begin{axis}[exppanel,
    yticklabel=\empty,   % y-axis shared with (a); frees the width for plot area
]
\addplot[fill=expbase]   coordinates {(Buchstabe,2.05)(Nummer,2.27)(Satz,2.28)(Absatz,1.36)(Paragraph,0.92)};
\addplot[fill=expcited]  coordinates {(Buchstabe,1.46)(Nummer,1.56)(Satz,1.54)(Absatz,0.96)(Paragraph,0.66)};
\addplot[fill=expparent] coordinates {(Buchstabe,0.67)(Nummer,0.69)(Satz,0.86)(Absatz,0.36)(Paragraph,0.92)};
\addplot[fill=expHybrid] coordinates {(Buchstabe,0.61)(Nummer,0.63)(Satz,0.75)(Absatz,0.34)(Paragraph,0.66)};

% Dashed baselines
([xshift=-7pt]axis cs:Buchstabe,2.05) -- ([xshift=7pt]axis cs:Buchstabe,2.05);
 ([xshift=-7pt]axis cs:Nummer,2.27)    -- ([xshift=7pt]axis cs:Nummer,2.27);
 ([xshift=-7pt]axis cs:Satz,2.28)      -- ([xshift=7pt]axis cs:Satz,2.28);
 ([xshift=-7pt]axis cs:Absatz,1.36)    -- ([xshift=7pt]axis cs:Absatz,1.36);
 ([xshift=-7pt]axis cs:Paragraph,0.92) -- ([xshift=7pt]axis cs:Paragraph,0.92);
% \draw[gray!75,dashed,line width=0.5pt]
\node[anchor=south, font=\tiny, text=expparent, inner sep=1pt]
      at ([xshift=2pt]axis cs:Absatz,0.36) {0.36};
\end{axis}
\end{tikzpicture}
% \vspace{-4pt}
\caption{charged: $W{=}$ $W_{\ell}^{m}$}
\label{fig:exp-charged}
\end{subfigure}

\caption{Impact of 1-hop context expansion on token efficiency (ToER) at top-5 retrieved chunks. Evaluation is conducted under matching at the \emph{Absatz} evaluation depth across all chunking granularities for unexpanded retrieval and three expansion strategies. Granularities are ordered from higher to lower, and the y-axis is shared between figures. \Cref{fig:exp-uncharged} is \emph{uncharged} ($W{=}W_{\ell}$): expanded context is added to retrieval but not considered against the token budget. \Cref{fig:exp-charged} is \emph{charged} ($W{=}W_{\ell}^{m}$): the expansion tokens count toward $W$, exposing the efficiency cost of each strategy.}
\label{fig:expansion-tokeneff}
\end{figure}

\section{Conclusion}
This paper addressed the challenge of verifiable retrieval in legal AI by introducing a structure-driven Knowledge Graph (KG) approach for German statutory law. By evaluating various chunking and one-hop context expansion strategies, we demonstrated that aligning retrieval mechanisms with the inherent hierarchical structure of legal codes is essential for maximizing both citation recall and token efficiency. 

\paragraph{Limitations} Even under the best configuration, the absolute $\text{ToER}$ is modest and decays steeply with evaluation depth. The strongest chunking granularities reach a $\text{ToER}$ of $\approx 14.2$ at the \textit{Code} evaluation depth, but only $\approx 3.2$ at the \textit{\S{}} and $\approx 1.5$ at the \textit{Satz} evaluation levels. This decay reflects the inherent difficulty of the task. This difficulty arises due to the domain gap between factual case descriptions (\textit{Tatbestand}) and abstract legal norms. Consequently, dense retrieval embeddings struggle to bridge non-legal narrative language to precise statutory formulations without explicit domain adaptation.

\paragraph{Future Work} The results of this paper can be expanded upon in multiple ways. First, a comparison between multiple embedders would provide valuable insight into the correlation between embedding strategy and the effectiveness of the KG-driven chunking approach. Second, structure-driven KG could benefit from the addition of entity-driven KG, which would encode additional relevance links within the document. Third, as different chunking levels utilize different proportions of context windows, additional experiments into comparative token efficiency with varied top-\textit{k} retrieval between KG node levels are a valuable avenue of future research. Furthermore, while granular chunking improves retrieval efficiency, its direct effect on downstream generation quality and hallucination mitigation remains to be validated, given that this study explicitly isolates retrieval performance from downstream LLM artifacts. Lastly, evaluation of retrieved items by legal specialists with relevant domain knowledge would provide valuable insight into their sufficiency for informed legal judgment by a human specialist.

\paragraph{Data and code availability.}
The knowledge graph, the six chunk corpora, the evaluation set (BGH case
identifiers with parsed ground-truth citations and a reconstruction script for
the decision text), and all evaluation results are archived on Zenodo at
\url{https://doi.org/10.5281/zenodo.22110527}. The reference implementation of
the metric and the retrieval/evaluation pipeline is at \url{https://github.com/ibrahimssd/structure-driven-legal-kg-rag}.

\section*{Declaration on Generative AI}
During the preparation of this work, the authors used Claude, Gemini, and Quillbot AI for the following purposes: grammar and spelling check, paraphrasing and rewording, and generating images through generating LaTeX code for Figure 1 to demonstrate non-generated data. After using these tools, the authors reviewed and edited the content as needed and take full responsibility for the publication’s content.

\bibliography{rage-kg-2026.bib}

@inproceedings{ostendorff2020towards,
  title={Towards an open platform for legal information},
  author={Ostendorff, Malte and Blume, Till and Ostendorff, Saskia},
  booktitle={Proceedings of the ACM/IEEE Joint Conference on Digital Libraries in 2020},
  pages={385--388},
  year={2020}
}

@incollection{darji2025segmentation,
  title={Segmentation and Processing of German Court Decisions from Open Legal Data},
  author={Darji, Harshil and Heckelmann, Martin and Kratsch, Christina and de Melo, Gerard},
  booktitle={Legal Knowledge and Information Systems},
  pages={276--281},
  year={2025},
  publisher={IOS Press}
}

@article{darji2026bundesrecht,
  title={Bundesrecht: An Open Library and Corpus for German Statutory Reference Processing},
  author={Darji, Harshil and Heckelmann, Martin and Kratsch, Christina and de Melo, Gerard},
  journal={arXiv preprint arXiv:2605.31338},
  year={2026}
}

@article{de2025graph,
  title={Graph rag for legal norms: A hierarchical and temporal approach},
  author={de Martim, Hudson},
  journal={arXiv e-prints},
  pages={arXiv--2505},
  year={2025}
}

@inproceedings{nigam2025nyayarag,
  title={Nyayarag: Realistic legal judgment prediction with rag under the indian common law system},
  author={Nigam, Shubham Kumar and Patnaik, Balaramamahanthi Deepak and Mishra, Shivam and Thomas, Ajay Varghese and Shallum, Noel and Ghosh, Kripabandhu and Bhattacharya, Arnab},
  booktitle={Proceedings of the 14th International Joint Conference on Natural Language Processing and the 4th Conference of the Asia-Pacific Chapter of the Association for Computational Linguistics},
  pages={1709--1726},
  year={2025}
}

@inproceedings{ongris2025benchmarking,
  title={Benchmarking KG-based RAG Systems: A Case Study of Legal Documents},
  author={Ongris, Jaycent G and Darari, Fariz and Tobing, Berty CL and Faisal, Douglas R and Lee, On},
  booktitle={CEUR Workshop Proceedings},
  volume={4079},
  pages={69--82},
  year={2025},
  organization={CEUR-WS}
}

@article{chen2026legalcitebench,
  title={LegalCiteBench: Evaluating Citation Reliability in Legal Language Models},
  author={Chen, Sijia and Yin, Hang and Zhou, Shunfan},
  journal={arXiv preprint arXiv:2605.10186},
  year={2026}
}

@article{guha2023legalbench,
  title={Legalbench: A collaboratively built benchmark for measuring legal reasoning in large language models},
  author={Guha, Neel and Nyarko, Julian and Ho, Daniel and R{\'e}, Christopher and Chilton, Adam and Chohlas-Wood, Alex and Peters, Austin and Waldon, Brandon and Rockmore, Daniel and Zambrano, Diego and others},
  journal={Advances in neural information processing systems},
  volume={36},
  pages={44123--44279},
  year={2023}
}

@InProceedings{Siddig2025_1260,
   author = {Ibrahim Siddig and Sviatoslav Tugeev and Munir Georges},
   booktitle = {Studientexte zur Sprachkommunikation: Elektronische Sprachsignalverarbeitung 2025},
   title = {Pattern-based parsing of German traffic regulations (StVO) for legal knowledge graph construction},
   year = {2025},
   editor = {Sven Grawunder},
   month = mar,
   pages = {255--264},
   publisher = {TUDpress, Dresden},
   isbn = {978-3-95908-803-9},
   issn = {0940-6832},
   url = {https://www.essv.de/pdf/2025_255_264.pdf},
}

@article{liu2024lost,
  title={Lost in the middle: How language models use long contexts},
  author={Liu, Nelson F and Lin, Kevin and Hewitt, John and Paranjape, Ashwin and Bevilacqua, Michele and Petroni, Fabio and Liang, Percy},
  journal={Transactions of the association for computational linguistics},
  volume={12},
  pages={157--173},
  year={2024}
}

@inproceedings{lu2026hichunk,
  title={HiChunk: Evaluating and Enhancing Retrieval Augmented Generation with Hierarchical Chunking},
  author={Lu, Wensheng and Chen, Keyu and Shen, Zhifeng and Qiao, Ruizhi and Sun, Xing},
  booktitle={Proceedings of the 64th Annual Meeting of the Association for Computational Linguistics (Volume 1: Long Papers)},
  pages={29738--29753},
  year={2026}
}

@article{xiao2025terag,
  title={TERAG: Token-Efficient Graph-Based Retrieval-Augmented Generation},
  author={Xiao, Qiao and Tsang, Hong Ting and Bai, Jiaxin},
  journal={arXiv preprint arXiv:2509.18667},
  year={2025}
}

@article{gutierrez2024hipporag,
  title={Hipporag: Neurobiologically inspired long-term memory for large language models},
  author={Guti{\'e}rrez, Bernal J and Shu, Yiheng and Gu, Yu and Yasunaga, Michihiro and Su, Yu},
  journal={Advances in neural information processing systems},
  volume={37},
  pages={59532--59569},
  year={2024}
}

@article{zhao2025kalm,
  title={Kalm-embedding-v2: Superior training techniques and data inspire A versatile embedding model},
  author={Zhao, Xinping and Hu, Xinshuo and Shan, Zifei and Huang, Shouzheng and Zhou, Yao and Zhang, Xin and Sun, Zetian and Liu, Zhenyu and Li, Dongfang and Wei, Xinyuan and others},
  journal={arXiv preprint arXiv:2506.20923},
  year={2025}
}

@article{ji2023survey,
    author={Ji, Ziwei and Lee, Nayeon and Frieske, Rita and Yu, Tiezheng and Su, Dan and Xu, Yan and Ishii, Yejin and Bang, Yejin and Marone, Andrea and Madotto, Andrea and Fung, Pascale},
    title={Survey of Hallucination in Natural Language Generation},
    journal={ACM Computing Surveys},
    volume={55},
    number={12},
    pages={1--38},
    year={2023},
    publisher={ACM}
    }

\appendix

\end{document}